\pdfoutput=1

\documentclass[11pt]{article}

\usepackage[]{ACL2023}

\usepackage{times}
\usepackage{latexsym}

\usepackage[T1]{fontenc}

\usepackage[utf8]{inputenc}

\usepackage{microtype}

\usepackage{inconsolata}

\usepackage{url}
\usepackage{amsthm}
\usepackage{amsmath}
\usepackage{centernot}
\usepackage{multirow}
\usepackage{comment}
\usepackage{enumitem}
\usepackage{graphicx}

\usepackage{amssymb}
\usepackage{pifont}

\newtheorem{myobs}{Observation}
\newtheorem{mytheorem}{Theorem}

\newcommand{\nearsynonym}{synonym}

\newcommand{\B}{\textbf}
\newcommand{\ften}{\texttt{F10}}
\newcommand{\fflc}{\texttt{FFLC}}
\newcommand{\ffsc}{\texttt{FFSC}}
\newcommand{\osptsl}{\texttt{OSPT-SL}}
\newcommand{\osptml}{\texttt{OSPT-ML}}

\usepackage{algorithm}

\usepackage{algpseudocode}
\algtext*{EndWhile}
\algtext*{EndIf}
\algtext*{EndFor}
\algtext*{EndFunction}

\algblock{Input}{EndInput}
\algnotext{EndInput}
\algblock{Output}{EndOutput}
\algnotext{EndOutput}

\algblock{Produces}{EndProduces}
\algnotext{EndProduces}

\date{}

\title{One Sense per Translation}
\author{Bradley Hauer, Grzegorz Kondrak \\
  Alberta Machine Intelligence Institute \\
  Department of Computing Science \\ 
  University of Alberta, Edmonton, Canada \\
  \texttt{{\{bmhauer,gkondrak\}@ualberta.ca}} \\
}

\begin{document}
\maketitle

\begin{abstract}
Word sense disambiguation (WSD) is the task
of determining the sense 
of a word in context.
Translations have been used in WSD as a source of knowledge,
and even as a means of delimiting word senses.
In this paper,
we define three theoretical properties 
of the relationship
between senses and translations, 
and argue that they constitute necessary conditions for using translations as sense inventories. 
The key property of 
One Sense per Translation (OSPT) 
provides a foundation for 
a translation-based WSD method.
%
The results of an intrinsic evaluation experiment 
indicate that our method achieves a precision of approximately 93\%
compared to manual corpus annotations. 
Our extrinsic evaluation experiments
demonstrate WSD improvements of up to 4.6\% F1-score 
on difficult WSD datasets.
\end{abstract}

\section{Introduction}
\label{intro}

Word sense disambiguation (WSD) is the task 
of classifying a word in context according to its 
sense. 
For example, given the context 
``the \underline{field} was covered in green grass,''
a WSD system would need to classify \emph{field} 
as having its ``flat open land'' sense,
rather than its ``area of study'' sense.
Throughout its history, 
WSD has been associated with 
translation \cite{weaver1949},
as it is understood that different senses of a word may translate differently.
%
For instance, in the above example, 
\emph{field} could be translated into French as \emph{champ},
but not as \emph{domaine}
(the latter could, however, translate the ``area of study'' sense of \emph{field}). 
In this paper, we address the open question:
\emph{to what extent can a translation-based method improve modern WSD?}

This question is surely an important one:
WSD remains an active area of research
\cite{blevins2020,barba2021escher,barba2021consec},
but despite the rapid improvements 
brought on by transformer-based \cite{vaswani2017} language models
such as BERT \cite{devlin2019},
substantial room for improvement remains \cite{maru2022}.
WSD has been used as a benchmark to compare and analyze 
transformer-based language models \cite{loureiro2021}.
It has also been shown to have applications to tasks such as
translation \cite{liu2018},
semantic parsing \cite{martinez2022},
and
metaphor detection \cite{maudslay2022}.
New variants of the task are still being proposed, 
such as visual WSD, in which candidate senses are represented by images
\cite{raganato2023vwsd}.
%
Clearly, the ability to map a word in context 
to an entry in a discrete lexical knowledge base
remains relevant in natural language processing,
for both human end users and downstream tasks.

Incorporation of translation information has been shown to be useful 
for both classic \cite{dagan1991} and modern \cite{luan2020} WSD methods.
%
%
%
Despite such proof-of-concept works,
current state-of-the-art WSD methods do not explicitly leverage translation,
leaving a potential source of knowledge untapped.
It is therefore of interest to the lexical semantics community 
to investigate the extent to which senses and translations correspond,
and how this correspondence can be leveraged in practice.

Our investigation has the following structure: \@
(1)~We begin by clearly defining the theoretically ``ideal''
mapping between senses and translations.
(2)~We show that such mappings are rare 
in practice, 
even between unrelated languages, 
offering an explanation as to why translation-based WSD methods 
became less common as the field developed. 
(3)~We posit 
that it is possible 
to improve supervised WSD performance by
leveraging instances where the translation of a word {\em does} 
determine its sense.
(4)~We propose and evaluate a 
translation-based disambiguation method 
to test this hypothesis.
(5)~We discuss the relationship 
between various theoretical properties 
and synonymy and polysemy.

Our empirical results strongly support our hypothesis.
A large-scale intrinsic evaluation 
of our method 
using existing lexical knowledge bases
shows that it achieves very high precision.
%
Our extrinsic evaluation 
shows that 
synthetic training data
produced by our method,
when used to train a supervised model,
can yield improvements in F1-score of up to 4.6\% 
on difficult WSD benchmark datasets.
%
We conclude that the explicit incorporation of 
contextual translations has great potential
to improve WSD research, and lexical semantics research in general.


The principal goal of our paper is 
the examination of the sense-translation connection from both theoretical and empirical perspectives in a modern context.
%
Thus our contributions 
are twofold:
a theoretical analysis of the relationship between senses and translation,  
supported by empirical analysis; 
and a method for efficient, unsupervised, 
large-scale semantic annotation via translations, 
which yields substantial WSD improvements. 

\section{Related Work}
\label{relwork}

%
%

The use of translations as a source of information about word senses
rose to prominence in the 1990s,
supported by the increasing availability 
of machine-readable multilingual resources.
\newcite{brown1991} and \newcite{dagan1991} 
developed statistical approaches to WSD,
with the former presenting a direct application 
to statistical machine translation.
%
%
\newcite{gale1992b} were the first to explicitly define WSD
in terms of identifying the correct translation:
they identify a set of six English words, each with two senses,
with a one-to-one mapping between those senses and their French translations.
%
%
This paradigm of translation-informed WSD
influenced the landmark WSD works of 
\newcite{yarowsky1995} and \newcite{schutze1998}, among others.
By the late 1990s, translation was so prevalent in the WSD literature
that \newcite{resnikyarowsky1997} explicitly proposed
``to restrict a word sense inventory to those distinctions 
that are typically lexicalized cross-linguistically.''

Interest in translation in the WSD literature continued throughout the 2000s
\cite{ide2000,chan2007,apidianaki2008},
culminating in two SemEval-2010 shared tasks:
cross-lingual lexical substitution
\cite{mihalcea2010semeval},
and cross-lingual WSD 
\cite{lefever2010semeval}.
The former can be viewed as 
the task of finding translations for a word in a given context.
%
In the latter,
translations from word-aligned parallel corpora were used to create
a ``multilingual sense inventory''.
The dataset was limited to small lexical samples,
and involved substantial manual-annotation effort for each tested language pair.
Neither the exact annotation criteria
nor the datasets themselves are available.
%
%

\newcite{yao2012} observed that prior work 
made conflicting assumptions 
about the correspondence between senses and translations.
They consider the case where a single word $e$ in a parallel corpus is aligned,
in different contexts,
with two different words, $f_1$ and $f_2$, in another language.
They point out that
some prior works, 
such as \newcite{lefever2011},
assume that $e$ is polysemous,
with $f_1$ and $f_2$ translating distinct senses of $e$,
while others,
such as \newcite{bannard2005},
instead assume that $f_1$ and $f_2$ translate a single sense of $e$,
and so are synonymous.
Our work builds upon this observation,
analyzing the various possible relations between senses and translations
in greater detail, 
and leveraging them them 
to improve WSD.

Despite the early successes of translation-based WSD,
methods based on monolingual resources,
namely WordNet \cite{miller1990lexi} and SemCor \cite{miller1993},
became prominent in the 2010s.
%
\emph{It Makes Sense} \cite{zhong2010},
a supervised WSD system based entirely on monolingual contextual features,
remained state-of-the-art for most of the decade \cite{papandrea2017}
%
%
before being replaced by 
methods based on contextual embeddings
\cite{hadiwinoto2019}.
%
In the early 2020s, 
WSD systems leveraging increasingly sophisticated
pre-trained language models
approached and finally exceeded 80\% accuracy on standard WSD datasets
\cite{blevins2020,barba2021escher,barba2021consec}.
In response to these advances,
\newcite{maru2022} proposed to focus on more difficult WSD instances,
such as those involving rare senses,
or on which modern WSD systems tend to make errors.
We support this proposal, 
and make use of their ``challenge'' datasets
in our experiments.

\section{Mapping Senses and Translations}
\label{theory}

While the use of translation information 
to identify
or 
even {\em define}
word senses was frequent in early WSD research, 
today it primarily
serves as supplementary data,
rather than as the core of the method \cite{luan2020}.
In this section, we lay the theoretical groundwork 
for explaining this paradigm shift;
an empirical analysis
follows in the next section.


Given an ideal one-to-one mapping 
between senses of a word and its lexical translations, 
each sense could be unambiguously defined by a distinct translation, 
and each translation would indicate a different sense. 
%
Figure \ref{fig:mapping} shows 
a graphical representation of a sense-translation mapping
which does \emph{not} conform to this ideal,
with three Italian translations 
of the English noun \emph{wood}. 
An edge between a sense and a translation indicates that 
the former can be translated by the latter.
As the sense-translation mapping is not bijective,
we cannot use translation knowledge alone
to determine the sense of an instance of \emph{wood}.


We can analyze the theoretical properties of such a mapping in terms of 
three word-level binary predicates, 
which are defined
on a given source word $e$
and language of translation $F$.
Each of these predicates is a necessary condition 
for such an ideal mapping to exist.
Moreover, 
in conjunction,
they represent a sufficient condition for 
using a word's translations as a sense inventory. 
The three {\rm sense-translation mapping} predicates are 
discussed in the following subsections.

\subsection{One Sense per Translation (OSPT)}
\label{ospt}

One Sense per Translation (OSPT) is 
the key predicate for translation-based WSD,
as it facilitates the inference of a word's sense from its translation.
OSPT underlies the method that we propose in Section \ref{methods}.

{\bf \mbox{OSPT}$(e,F) := $} 
``all senses of the word $e$ 
have disjoint sets of lexical translations in language $F$''

\begin{figure}[t]
\includegraphics[width=\linewidth]{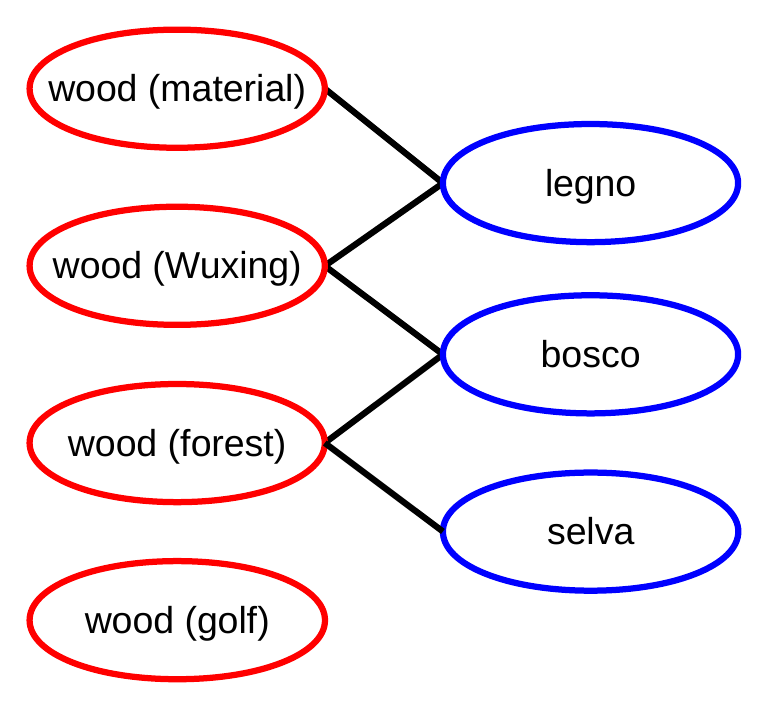}
\caption{
An example mapping between 
senses and translations.
Each translation corresponds to at least one sense.
}
\label{fig:mapping}
\end{figure}

If OSPT holds,
each translation of $e$ corresponds to exactly one sense, 
and so we can use the sense-translation mapping to perform WSD.
%
Exceptions to OSPT 
occur when words from different languages 
share multiple senses, 
a phenomenon which we refer to as 
{\em parallel polysemy}.
For OSPT to hold, 
the source word cannot exhibit parallel polysemy with any of its translations.
For example, 
Figure~\ref{fig:mapping} shows a violation of OSPT,
as the Italian word {\em legno}
maps to two distinct senses of the English word \emph{wood}.
Therefore, 
the sense of \emph{wood} in a given sentence  
cannot be inferred solely from the fact that
it is translated as \emph{legno}.
On the other hand, if an instance of 
\emph{wood} is translated into Italian as \emph{selva}, 
we can infer that it is used in its ``forest'' sense.


\subsection{One Translation per Sense (OTPS)}

The One Translation per Sense (OTPS) 
predicate can be viewed as a dual of OSPT,
reversing the roles of senses and translations.

{\bf \mbox{OTPS}$(e,F) := $} 
``each sense of the word $e$ 
has at most one lexical
translation in language $F$''

In other words, no pair of translations translate the same sense.
Exceptions to OTPS are instances of synonymy between translations of a given source
word.\footnote{Interestingly, the WSD algorithm of \newcite{diab2002},
which disambiguates English words based on their French translations,
is based on the assumption that all target-language words are monosemous.}
%
For example, 
the ``forest'' sense in Figure \ref{fig:mapping}
maps to two distinct translations,
\emph{bosco} and \emph{selva},
violating OTPS.
This presents a challenge to the proposal 
to use translations as sense inventories
\cite{resnikyarowsky1997}
by creating cases where instances of a word 
need not be
distinguished by their translations.
Moreover, 
this also poses a problem for aligning sense distinctions with translation distinctions
\cite{lefever2010semeval},
as \emph{bosco} and \emph{selva} must somehow be ``clustered''
to avoid identifying instances of \emph{wood} with these translations
as being semantically distinct.
%
%
Note, however, that unlike violations of OSPT, 
translations that cause OTPS violations 
can still be used to disambiguate
the translated word in some cases.

\subsection{No Lexical Gaps (NoLG)}

The No Lexical Gaps (NoLG) predicate
reflects the importance  
of {\em lexical gaps}
\cite{bentivogli2000} in multilingual semantics.

{\bf \mbox{NoLG}$(e,F) := $} 
``each sense of the word $e$ 
has at least one translation in language $F$''

Since it is not practical 
to enumerate all possible phrasal translations of each sense,
such lexical gaps generally preclude translation-based WSD:
we cannot identify a sense based on its lexical translation
if it \emph{doesn't have} a lexical translation.
For example,
the ``golf'' sense of \emph{wood} in Figure~\ref{fig:mapping}
corresponds to a lexical gap, 
and so would need to be translated
into Italian by a compositional phrase, such as ``legno da golf''.
%

In summary, 
an ideal one-to-one sense-translation mapping
seems to be a very brittle structure.
Any exception to OSPT, OTPS, or NoLG would complicate 
the use of translations to define sense inventories.
Moreover, any exception to OSPT or NoLG will outright preclude
the use of translations alone for WSD.
The viability of translation-informed WSD
therefore rests on the extent to which these properties hold in practice, 
which we investigate in the next section. 

\section{Empirical Analysis}
\label{experiments}






%
%

We focus on English, with three languages of translation
which represent various degrees of relatedness to English:
Italian, Polish, and Chinese,
For each language,
we compute the proportion of English words 
for which OSPT, OTPS, and NoLG hold
in BabelNet \cite{navigli2012},
a large lexical knowledge base frequently used as a sense inventory for multilingual WSD \cite{pasini2021xlwsd}.
We consider only English words
with at least two senses in WordNet 3.0
(BabelNet inherits senses from WordNet),
and at least one translation in the target language in BabelNet 4.0.
There are 
20,426 such words with Italian as the target language,
17,404 for Polish,
and 19,973 for Chinese.

Table \ref{table_assumptions} summarizes the results.
The NoLG values indicate that
the majority of English words involve at least one lexical gap in any of the three languages of translation. 
The OTPS row shows that
even fewer words have no more than one translation per sense. 
The OSPT property is more reliable,
covering almost 60\% words with Italian as the language of translation, 
and approaching 80\% with less related languages such as Polish or Chinese.
However, 
the last row in the table 
demonstrates that 
only a very small percentage of English words 
satisfy all three properties at the same time.

Since we have argued that the conjunction of
the three properties is 
a necessary condition
for an ideal one-to-one sense-translation mapping,
these empirical results provide 
an explanation why using translations as sense inventories is infeasible in practice.
Furthermore, 
even if we had a sense inventory with a complete 
mapping between senses and translations
(something BabelNet and comparable resources aspire to provide),
the OSPT values in our results table indicate that 
a substantial portion of words 
cannot be disambiguated on the basis of their translations alone. 
We conclude that this was a key factor
in the abandonment of the use of translations 
to induce sense inventories, or perform WSD on all words.
Nevertheless, 
we posit that
translations {\em can} be leveraged to improve WSD,
specifically be exploiting those cases 
where a translation of a word in context 
uniquely determines its sense.
In the next section, we present and apply a 
method for using translations to tag a subset of the tokens in a parallel corpus.

\begin{table}[t]
\begin{small}
\centering
\renewcommand{\arraystretch}{1.5}
\begin{tabular}{|c|c|c|c|}
\hline
            & Italian & Polish  & Chinese \\\hline\hline
OSPT 
& 59.5 & 77.4 & 75.7 \\\hline
OTPS 
& 16.3 & 22.8 & 10.3 \\\hline
NoLG        & 47.4 & 38.3 & 40.3 \\\hline
ALL & 1.9 &  2.6 &  1.5 \\\hline
%
%
\end{tabular}
\caption{The percentage of English polysemous words in BabelNet
which exhibit each of the three sense-translation mapping properties
with respect to three languages of translation.}
\label{table_assumptions}
\end{small}
\end{table}

\section{Corpus Tagging with OSPT}
\label{methods}

Although the results in Section \ref{experiments} 
demonstrate that translations alone are not sufficient for all-words WSD,
prior work such as 
\newcite{gale1992b} and \newcite{lefever2010semeval}
have shown that they can still be applicable to lexical samples.
%
In this section, we explore the idea of 
using translations to improve WSD on modern standard datasets. 
Specifically, we leverage those cases where the translation of a word
corresponds to exactly one of its senses in order 
to create supplementary training data for a supervised WSD system.

\subsection{Corpus Tagging}

The generation of ``silver datasets'' for WSD 
is a way to address 
the knowledge acquisition bottleneck \cite{pasini2021},
the difficulty of obtaining training data for supervised WSD.
To this end,
the goal of semantic corpus tagging is not to disambiguate all word tokens,
or any particular subset of lemmas;
rather, the goal is to 
partially sense-annotate a corpus 
to produce supplementary training data for a supervised WSD system. 

Automatic sense tagging 
has been a popular area of research in lexical semantics.
\newcite{taghipour2015} used a mapping
of Chinese translations to English senses to annotate the English side
of an English-Chinese parallel corpus;
however, this mapping 
is not available.
\newcite{pasini2017} sense-tag Wikipedia articles
using a variant of the personalized page-rank algorithm (PPR),
while \newcite{dellibovi2017} applies a similar approach 
to the EuroParl parallel corpus.
\newcite{barba2020} use a pre-trained language model to identify
semantically-equivalent translations of manually sense-annotated tokens.
Most recently, \newcite{hauer2021annotation} 
propose a family of pipeline approaches employing WSD methods,
machine translation, lexical resources, and various filtering techniques.

Our work differs from prior work on using translations for WSD
in that
(a) we show that our method can achieve good results with only one language of translation, 
(b) our method is independent of statistical information such as relative sense frequencies, 
and (c) our method does not explicitly require any contextual information.
%
In contrast, 
the method of \newcite{apidianaki2015} 
backs off to the BabelNet first sense (BFS), a frequency-based baseline,
if it is unable to narrow down the sense of the target word. 
This back-off strategy is particularly undesirable for
tagging tokens that correspond to rare word senses.
Moreover, their method is tested only with multiple languages of translation,
and is applied directly to all-words WSD on a parallel corpus,
rather than to generation of high-precision training data.
%
The method of \newcite{bonansinga2016} similarly 
depends on sense frequency information, 
and is evaluated only intrinsically, 
with multiple languages of translation.
The method of \newcite{luan2020} depends on an existing disambiguation of the text, 
in addition to translations.
Thus, our method is unique 
in that it
can produce supplementary WSD training data with minimal assumptions about the available resources.

\subsection{Method}

\begin{figure}
\begin{algorithmic}
\For{each token $e$ on the $S$ side of $C$}
    \If{$\exists$ token $f$ aligned with $e$}
        \State $M_e \gets$ the set of synsets containing $e$
        \State $M_f \gets$ the set of synsets containing $f$
        \If{$|M_e \cap M_f| = 1$}
            \State Let $s$ be the sole synset in $M_e \cap M_f$
            \State Tag $e$ with sense $(e,s)$
        \EndIf
    \EndIf
\EndFor
\end{algorithmic}
\caption{Pseudo-code for the sense tagging algorithm.}
\label{alg-sense-tagging}
\end{figure}

%
Our 
method 
is inspired by \newcite{loureiro2020}.
They sense annotate only tokens that correspond to monosemous words,
i.e., those that have only one sense,
which is a trivial task in itself. 
%
However, they also show that 
a WSD method which propagates information
between senses of different words
can benefit from these annotations.
For example,
the monosemous word \emph{airplane} 
is a synonym of the word \emph{plane},
which is polysemous. 
Therefore, an annotated instance of \emph{airplane} can inform a model
about the context in which the corresponding sense of \emph{plane} may appear.
%
%

In our approach, 
instead of monosemous words,
we sense tag tokens which
can be disambiguated based on their translations.
%
For example, the English noun \emph{vault} has four senses,
corresponding to a burial vault, a bank vault, an arched ceiling,
or a jump over an obstacle.
%
The Polish 
word \emph{wolta} 
can translate only
the ``jump'' sense.
Therefore, if we find an instance of \emph{vault} translated as \emph{wolta},
we can annotate \emph{vault} with its ``jump'' sense,
as no other sense could have been so translated.
The absence of parallel polysemy between {\em vault} and {\em wolta} 
is a sufficient condition for the correctness of this annotation,
regardless of whether OSPT holds for all Polish translations of {\em vault}. 
Our method uses this approach to partially annotate a parallel corpus,
creating new sense-annotated WSD training data.
Our hypothesis is that 
adding our translation-based annotations to a standard training corpus
will improve the results of 
a supervised WSD system.

We follow the theoretical framework of \newcite{hauer2023set}.
The sense inventories, as well as the mapping between senses and translations,
can be obtained from 
a multilingual wordnet, such as BabelNet.
Multilingual wordnets consist of synonym sets, or \emph{synsets}, 
each corresponding to a concept,
and containing the words which can express that concept.
The synsets that contain a word correspond to its senses;
a sense can be viewed as a pair of a word and a synset that contains it.
The target-language words in that synset 
are the words which can translate that sense.
For example, in  Figure \ref{fig:mapping},
a multilingual wordnet should have 
a synset 
corresponding to the concept of ``wood (material)''
which contains \emph{wood} and \emph{legno},
but not \emph{selva}.

The pseudo-code of the algorithm is shown in Figure \ref{alg-sense-tagging}.
It takes as input a sentence-aligned parallel corpus $C$, 
involving the source language $S$ 
(in our experiments, English)
and the target language $T$,
which has been 
tokenized, lemmatized, POS-tagged, and word-aligned. 
The algorithm generates sense tags for a subset of the 
tokens on the source side of $C$. 
%
The algorithm consults a wordnet that covers
languages $S$ and $T$.
For each content word token $e$ on the source side
aligned with a single target-language token $f$,
we determine the number of synsets 
which contain both $e$ and $f$.
Since each sense of a word 
uniquely corresponds to a synset containing that word,
this is equivalent to determining how many senses of $e$
can be translated by $f$.
If the result is exactly one, we annotate $e$ with its sense
corresponding to the synset $s$ that it shares with $f$.
%
For example,  
if an instance of \emph{wood} is aligned with \emph{selva},
it is tagged with its ``forest'' sense, 
given that 
it is the only sense of \emph{wood} which \emph{selva} can translate.

Our method is unsupervised, efficient, 
scalable, and fully explainable.
Its running time scales linearly with the size of the corpus.  
%
%
%
The resources upon which it depends
are freely available for a wide variety of languages. 
These include the parallel corpora our method annotates, a multilingual wordnet, 
as well as  tools for tokenization, POS-tagging, and alignment.
It operates purely on the basis of contextual translation,
without the need for additional tools such as knowledge-based WSD systems
or contextual embeddings.

\subsection{Intrinsic Evaluation}
\label{experiments2}

We test our translation-based corpus-tagging 
method on the manual sense annotations in 
MultiSemCor \cite[MSC,][]{bentivogli2005},
a word-aligned sense-annotated bitext, which 
was created by manually translating SemCor \cite{miller1993}.
It is tokenized, POS-tagged, 
and word-aligned with a knowledge-based aligner.
%
There are 91,937 English word tokens in MSC
annotated with exactly one WordNet 1.6 sense, 
and aligned with a single Italian word.
%
We randomly select 10,000 of these tokens, 
and strip them of their sense annotations
to form our test set.

As our multilingual wordnet,
we use MultiWordNet \cite[MWN,][]{pianta2002} 
version 1.5.0.
MWN was created by expanding Princeton WordNet 1.6
by adding Italian translations,
as well as new synsets to cover English lexical gaps.
%
To mitigate the sense omission errors in MWN, 
we enrich it 
with 81,937 sense-translation pairs 
from MSC, 
excluding those which are in our 10k-token test set.
%
%

The results of the application or our method to 
the 10,000 annotated tokens in the test set
yield 
a coverage of 33.3\% and  
%
a precision of 
92.6\%, 
with 
the majority of errors  caused by missing translations in MWN.
%
%
Thus, 
our unsupervised method 
achieves higher precision
than contemporary 
supervised WSD systems 
on standard English WSD datasets 
\cite{barba2021consec}.
While these results are not directly comparable
due to the different test sets,
%
we interpret this as strong evidence 
for the efficacy and utility of our method
for generating high-quality WSD training data.

%
%

\subsection{Extrinsic Evaluation}
\label{extrins}


Having
demonstrated that our 
method
can accurately disambiguate a subset of the tokens in a corpus, 
%
%
%
in this section
we test whether sense-annotated data produced in this way
can be used to improve the performance of a supervised WSD system.
This is achieved by appending the data that our translation-based method produces
to SemCor, 
a standard training corpus for English WSD.
Note that 
no manual sense annotations exist for
the corpus that we annotate in these experiments;
we are creating novel sense-annotated data.

\subsubsection{Experimental Setup}
\label{extrins_setup}

Our parallel corpus is the English-Italian part
of the OpenSubtitles corpus \cite{lison2016},
which contains approximately 35M sentence pairs.
%
We  
tokenize, lemmatize, and POS-tag both sides of the corpus
with TreeTagger \cite{schmid2013}
using 
pre-trained models.\footnote{\url{https://cis.uni-muenchen.de/~schmid}}
%
We perform word alignment with BabAlign \cite{luan2020},
which refines the output of 
FastAlign \cite{dyer2013} 
by leveraging BabelNet as a source of lexical knowledge.

We again derive a sense-translation mapping from 
MultiWordNet, 
but this time without adding information 
from MultiSemCor. 
%
Since MultiWordNet is based on WordNet 1.6,
we map each sense annotation to its most probable WordNet 3.0 equivalent,
using a publicly available probabilistic 
mapping.\footnote{\url{http://www.lsi.upc.es/~nlp}}

As our supervised WSD system, 
we adopt the latest version of LMMS \cite{loureiro2022}, 
which exploits relations between senses derived from WordNet 
in order to share information across related senses.

\begin{table}[t]
\centering
\small
\begin{tabular}{|c|c|c|c|c|}
\hline
Corpus & Tokens & Senses & Lemmas \\\hline\hline
SemCor  & 226,036 & 33,316 & 22,899 \\\hline
\hline
\ften{} & 219,793 & 28,589 & 23,033 \\\hline
\ffsc{} & 117,646 & 16,818 & 15,329 \\\hline
\fflc{} &  90,616 & 13,147 & 12,406 \\\hline
\end{tabular}
\caption{Statistics on the sets of sense annotations generated using 
the three 
filtering procedures.}
\label{table_extrins_stats}
\end{table}

\subsubsection{Filtering Annotations}
\label{extrins_methods2}

Supervised WSD systems tend to exhibit a bias toward
senses which are more frequent in the training data \cite{loureiro2020lm}.
Therefore, even a set of perfectly correct sense annotations
may degrade the model's performance
if the sense frequency distribution in the newly produced data
diverges from that of the test data,
which is not known in advance.
%
We therefore 
filter the generated annotations to avoid 
greatly altering the sense frequency distribution
of SemCor.

Following the example of \newcite{loureiro2020},
we limit the number of annotated instances of each individual sense to 10, 
selected at random. 
This not only helps to prevent highly unbalanced sense frequency distributions,
but also reduces the training time on the generated corpora. 
%
We refer to this set of instances as \ften{}.
In order to focus on gaps in the coverage on SemCor,
we also test two additional filtering strategies that are applied to 
the annotations in \ften{}.
The first 
filters for lemma coverage (\fflc),
by removing all annotations for {\em lemmas} which appear in SemCor.
The second 
filters for sense coverage (\ffsc),
by removing all annotations for {\em senses} which appear in SemCor.
Therefore, 
the \fflc{} annotations 
are a subset of the \ffsc{} annotations, 
which in turn are a subset of the \ften{} annotations. 
%

\subsubsection{Datasets} 
\label{extrins_data}

We obtain baseline results by training LMMS on SemCor,
specifically the version provided by \newcite{raganato2017}.
To test our method, we train  
three additional LMMS models
which augment SemCor annotations with  
\ften{}, \ffsc{}, and \fflc{}, respectively.
%
The sizes of these generated supplementary datasets, and of SemCor itself,
are shown in Table \ref{table_extrins_stats}.

\begin{table}[t]
\centering
\small
\begin{tabular}{|c|c|c|c|c|c|}
\hline
Dataset    & Full &  MFS &  LFS &  ZSS &  ZSL \\\hline\hline
SE2        & 2,282 & 1,486 &   796 &   385 &  255 \\\hline
SE3        & 1,850 & 1,213 &   637 &   198 &  112 \\\hline
S07        &   455 &   250 &   205 &    53 &   20 \\\hline
S13        & 1,644 & 1,031 &   613 &   341 &  202 \\\hline
S15        & 1,022 &   623 &   399 &   204 &  103 \\\hline
ALL        & 7,253 & 4,603 & 2,650 & 1,181 &  692 \\\hline
\end{tabular}
\caption{Number of instances in each of the subsets of each dataset
and the concatenation of all five datasets.}
\label{table_extrins_datasets}
\end{table}

We evaluate our models 
on the standard WSD benchmark of \newcite[][``R17'']{raganato2017}.
In addition to providing the baseline SemCor training corpus,
R17 also contains five English WSD test sets 
created for five shared tasks:
Senseval-2   \cite[SE2,][]{edmonds2001},
Senseval-3   \cite[SE3,][]{snyder2004},
SemEval-2007 \cite[S07,][]{pradhan2007},
SemEval-2013 \cite[S13,][]{navigli2013},
and
SemEval-2015 \cite[S15,][]{moro2015}.
Following prior work, 
we use the S07 dataset 
to develop our method.
%
We also evaluate our models on the concatenation of all five datasets,
referred to as ALL\footnote{This includes S07, as is standard in the WSD literature.},
using the provided evaluation program;
since LMMS disambiguates all words,
the metrics
precision, recall, F1, and accuracy 
are all equal throughout these experiments.

Following \newcite{blevins2020},
we also evaluate our models on the following subsets of ALL:
most frequent sense (MFS), 
less frequent sense (LFS),
zero-shot senses (ZSS),
and zero-shot lemmas (ZSL).
MFS and LFS are disjoint, 
and their union is the complete dataset; 
ZSL is a subset of ZSS.
Table \ref{table_extrins_datasets} shows the size of each such subset.

Finally,
we also test our models on 
five new benchmark datasets 
of \newcite[][``M22'']{maru2022}:
challenge (42D), amended ALL (ALLa), 
amended S10 (S10a), hardEN(hEN), 
and  softEN (sEN). 
%

\subsubsection{Results} 
\label{extrins_results}

The results in Tables \ref{table_extrins_results_full} and \ref{table_extrins_results_subsets}
%
%
show that adding supplementary training data
created by our method 
generally increases WSD accuracy, 
especially on rare and unseen senses.
On the recently proposed 42D and hardEN challenge sets, 
we observe accuracy improvements of 4.6\% and 2.5\% respectively,
using 
the \ften{} filtering strategy.
This same approach 
yields improvements on LFS, ZSS, and ZSL partitions of the R17 ALL set,
demonstrating that our method makes models more robust 
against such instances.
%
We interpret these results as 
evidence for the efficacy and utility
of our translation-based corpus tagging method.

\begin{table*}[t]
\centering
\small
\begin{tabular}{|c|c|c|c|c|c|c|c|c|c|c|c|c|}
\hline
\multicolumn{1}{|c|}{Training Data}  &  \multicolumn{6}{c|}{R17}      &  \multicolumn{5}{c|}{M22}   \\
&     SE2  &  SE3  &  S07  &  S13  &  S15  & ALL &  42D  &  ALLa  &  S10a  &  hEN  &  sEN \\\hline
\multicolumn{1}{|c|}{SemCor (Baseline)} &  76.1  &  \B{73.9}  &  67.0  &  \B{75.2}  &  77.4  &  75.0  &  35.9  &  74.9  &  \B{77.3}  &  12.6  &  \B{78.0} \\
\hline
SemCor +                             
\ften\;\:  &  74.9  &  72.6  &  65.9  &  72.8  &  78.2  &  73.7  &  \B{40.5}  &  73.3  &  77.1  &  \B{15.1}  &  76.6 \\
SemCor + 
\fflc  &  \B{76.7}  &  \B{73.9}  &  \B{67.5}  &  75.0  &  77.5  &  \B{75.1}  &  34.9  &  \B{75.1}  &  76.6  &  13.4  &  77.9 \\
SemCor + 
\ffsc  &  76.2  &  72.3  &  66.8  &  73.5  &  \B{78.3}  &  74.3  &  38.4  &  74.1  &  76.3  &  14.7  &  77.1 \\
\hline
\end{tabular}
\caption{
F1-scores (in \%) on the 10 WSD test sets.
SE07 is the development set.
The best results are 
in bold.
}
\label{table_extrins_results_full}
\end{table*}

The results further suggest 
that filtering generated annotations
has a substantial impact on 
the resulting model.
%
The frequency with which a word can be tagged with a particular sense
by leveraging lexical translation
need not correlate with the frequency of that sense in practice.
Therefore, when using such generated corpora,
care should be taken to select an appropriate filtering strategy.
For instance, in a corpus where unseen senses or words are expected
(e.g., 
in an unusual genre or domain),
the \ffsc{} filtering strategy may be  the best option,
as shown by its 
accuracy yields on ZSS and ZSL instances.

We conclude that
our method for translation-based sense tagging
offers substantial benefits, 
especially on difficult instances
\cite{blevins2021}. 
These improvements are 
obtained using a recent WSD method which is based on 
pre-trained transformer-based language models. 
This demonstrates that lexical translation can be a useful source of information even for modern WSD systems.

\begin{table}[t]
\centering
\small
\begin{tabular}{|c|c|c|c|c|c|}
\hline
\multicolumn{1}{|c|}{Training Data}  &  \multicolumn{4}{c|}{R17 - ALL}      \\
&      MFS   &  LFS   &  ZSS   &  ZSL   \\\hline
\multicolumn{1}{|c|}{SemCor (Baseline)} &    85.4  &  51.2  &  58.9  &  88.9  \\
\hline
SemCor +
                            \ften\;\: &    83.1  &  \B{52.1}  &  61.7  &  89.5  \\
SemCor +

                            \fflc  &    \B{85.5}  &  51.3  &  60.1  &  89.6  \\
SemCor +

                            \ffsc  &    83.9  &  51.9  &  \B{62.7}  &  \B{89.7}  \\
\hline
\end{tabular}
\caption{
F1-scores (in \%) on subsets of the concatenation of all R17 datasets.
The best results are in bold.
}
\label{table_extrins_results_subsets}
\end{table}

As a final note, we note that since
the phenomenon of {\em parallel polysemy} is closely related to 
that of {\em parallel homonymy},
our approach is well-suited to homonym-level disambiguation.
\newcite{hauer2020ohpt} 
argue that homonym distinctions are the coarsest
possible sense inventory, and that almost all homonyms
have disjoint sets of translations. 
Therefore, unlike OSPT, 
One Homonym per Translation (OHPT) \emph{does} hold in general. 
Our translation-based approach could therefore be applied 
with near-perfect accuracy
to disambiguate 
words 
at the homonym level.

\section{Discussion}

Our theoretical analysis 
in Section \ref{theory}
%
established that
OSPT is a sufficient condition 
for the ability to determine the sense of a word
given its translation in context.
However, 
the subsequent empirical analysis in Section~\ref{experiments} showed 
that OSPT does not hold in general.
%
Nevertheless, our experiments in Section~\ref{methods}
provide clear evidence that 
we can leverage translations to produce high-precision sense annotations
on the subset of word instances for which OSPT holds.
%
These results demonstrate the importance
of investigating the relations between 
senses, synonymy, polysemy, and translation.
In this section, we further explore these ideas,
taking the assumptions examined by \newcite{yao2012} 
(c.f., Section~\ref{relwork})
to their logical extremes.

\subsection{One Concept per Word: No Polysemy} 
\label{ocpw}


First, let us consider 
a hypothetical language in which polysemy does not exist; 
that is, every content word has exactly one sense.
In such a language, 
there could be no 
semantic ambiguity,
and so WSD would be trivial:
any given word could only express a single concept,
regardless of its context.
%
%
OSPT would always hold in such a language,
no matter the language of translation,
since each translation of a word could only translate its single sense.

To the best of our knowledge,
no natural language 
contains only monosemous words. 
For example, 
77.8\% of English words in BabelNet 
occur in only one synset, 
with many of those being rare or technical terms.
Similarly,
\newcite{loureiro2020} observe that
nearly 80\% of lemmas in WordNet have only one sense, 
which allows them 
to generate useful resources for WSD.
Only some constructed languages, such as Lojban/Loglan,
strive to enforce complete monosemy on the lexicon \cite{cowan1997}.

The untenable position that rejects any partitioning of word meanings into senses (``one sense per word'')
relates to various approaches to both theoretical and computational linguistics. 
%
In theoretical linguistics, the {\em monosemist} approach 
holds that different observed senses of a polysemous word result
from a combination of its unique core meaning with the pragmatics of each specific context
\cite{francois2008}.
In computational linguistics, methods that rely on 
exclusive use of static word embeddings, such as
those learned by word2vec \cite{mikolov2013}
make no allowance for discrete senses or sense embeddings. 

\subsection{One Word per Concept: No Synonymy}
\label{owpc}

Now, let us consider the opposite extreme:
a hypothetical language without synonymy.
%
If a wordnet were constructed for such a language,
every synset would contain exactly one word.
For any given concept, there would be at most one word
that could be used to express it.
%
One Translation per Sense (OTPS)
would always hold if such a language 
was used as the language of translation. 
%

Again, 
it is unlikely that the entire lexicon of any
natural language
could satisfy this requirement.
A language could perhaps be {\em constructed} according to this principle:
for example, in Esperanto, 
synonymy and homonymy are considered undesirable \cite{puvskar2015}.
%
Moreover, there will be a subset of any language 
which {\em does} satisfy this property.
Indeed, 
approximately 56\% of WordNet 3.0 synsets
contain only one word
(e.g., {\em proton}).

A similar position in computational linguistics
(``one sense per context'')
is diametrically opposite to the {\rm monosemist} 
approach described above. 
%
For example, \newcite{martelli2021} propose 
``dropping the requirement of a fixed sense inventory''
and instead using representations which assign 
each word token a unique contextualized embedding.
%
Such a position 
can be interpreted as an assignment of a unique sense 
to every occurrence of a given word in a distinct context.
In view of our theoretical investigation, 
such an approach is effectively incompatible with our definition of synonymy. 
Nevertheless, the existence of synonymy in any human language is widely accepted in linguistics. 
In addition, computational linguistics tasks, such as machine translation, need to account for synonymy, 
given that the goal is to produce fully fluent, 
rather than just semantically correct texts and utterances. 


\subsection{One Word $\equiv$ One Concept}
\label{interlingua}

If the two constraints described above are combined, 
it would result in 
a language that has neither polysemy nor synonymy. 
We refer to this hypothetical language as {\em Interlingua}.
In Interlingua,
every concept could be expressed by exactly one word,
which could express only that concept;
every synset would have a size of one,
and every word would be in one synset.
Assuming a sense-translation mapping is available,
e.g. via a multilingual wordnet which includes Interlingua,
lexical translation {\em into} Interlingua 
could be reduced to identifying the sense of 
the source word. 
The converse also holds: the sense of a word could always be identified,
given its translation into Interlingua.
Working in the other direction, 
given a perfect multilingual wordnet, 
finding a translation for an Interlingua word
would only require selecting a word from the corresponding synset
in the target language. 

Perhaps the most direct application for Interlingua 
is language-independent semantic parsing.
\newcite{martinez2022} propose the {\em BabelNet Meaning Representation} (BMR),
a semantic parsing formalism which converts an input sentence
into a language-independent representation.
Each content word is mapped to the unique identifier of the BabelNet synset
corresponding to the concept it refers to.
This creates a formal meta-language 
in which every concept is unambiguously expressed in exactly one way: 
by the corresponding BabelNet synset ID.
Hence, the BMR satisfies 
one ``word'' per concept \emph{and} one concept per ``word'',
with BabelNet IDs taking the place of words.
There is no synonymy, 
as each ID is by design unique in representing its particular concept,
nor is there polysemy,
as each ID is 
unambiguous in its reference
to some lexicalized concept. 
Thus, what may appear as a completely hypothetical and abstract construct 
can in fact be viewed as a theoretical model of a modern semantic approach. 

\section{Conclusion}
\label{conclusion}

In this paper, 
we formulate several propositions
related to senses, translations, synonymy, and polysemy.
We show empirically that
the assumptions that would allow translations to serve as a sense inventory
hold simultaneously only for a small fraction of words.
%
Nevertheless, 
we also demonstrate that 
the link between word senses and translations
is not merely of theoretical interest. 
%
In particular, we present a 
method for leveraging translations  
to perform high-precision unsupervised sense annotation.
%
We observe substantial WSD improvements
especially
on senses or lemmas that are less frequent or
not found at all in existing training data.

%
%
%
%
%
%
Considering the above applications 
to constructed languages, contextual embeddings, and semantic parsing, 
we intend  
to continue our theoretical investigations into
open issues in multilingual lexical semantics,
%
%
%
%
%
%
and guide empirical research toward
more explainable models and results.

\section*{Limitations}

The principal limitation of our sense-tagging method 
is its dependence on linguistic resources,
particularly text corpora and multilingual wordnets.
As is the case with any method which depends on such resources, 
the reliability of our method 
will vary 
depending on the language to which it is applied, and the quality of the resources available. 
Care should be taken when applying our method to languages and domains, 
where resources are limited in terms of availability, coverage, or correctness.
%
%
Any biases in these resources, 
e.g. biases toward English, 
may be inherited by our method. 
Likewise, the quality of translation and word alignment methods 
for pairs of languages will have a substantial impact 
on the quality of the data our method produces. 
Thus, before applying our method, 
we recommend assessing the quality of semantic resource coverage 
and translation and alignment quality 
for the languages under consideration.
Nevertheless, the state of resource coverage and quality within NLP is improving, 
and we expect the applicability of our method to improve concordantly.

%
%

\section*{Acknowledgements}

This research was supported by the Natural Sciences and Engineering Research Council of Canada (NSERC), and the Alberta Machine Intelligence Institute (Amii).
We also thank Robert Holte for providing additional comments.

\bibliography{wsd}
\bibliographystyle{acl_natbib}

\end{document}